\documentclass[11pt,a4paper]{article}
\usepackage[hyperref]{ranlp2023}
\usepackage{times}
\usepackage{latexsym}
\usepackage{graphicx}

\usepackage{microtype}

\aclfinalcopy 

\title{Can Model Fusing Help Transformers in Long Document Classification? An Empirical Study}

\author{Damith Premasiri$^\diamondsuit$, \textbf{Tharindu Ranasinghe$^\heartsuit$ and Ruslan Mitkov$^\spadesuit$} \\
 $^\diamondsuit$University of Wolverhampton, Wolverhampton, UK \\
 $^\heartsuit$Aston University, Birmingham, UK \\
 $^\spadesuit$Lancaster University, Lancaster, UK \\
 {\tt damith.premasiri@wlv.ac.uk}, 
 {\tt  t.ranasinghe@aston.ac.uk} \\ {\tt r.mitkov@lancaster.ac.uk} }

\date{}

\begin{document}
\maketitle
\begin{abstract}
Text classification is an area of research which has been studied over the years in Natural Language Processing (NLP). Adapting NLP to multiple domains has introduced many new challenges for text classification and one of them is long document classification. While state-of-the-art transformer models provide excellent results in text classification, most of them have limitations in the maximum sequence length of the input sequence. The majority of the transformer models are limited to 512 tokens, and therefore, they struggle with long document classification problems. In this research, we explore on employing \textit{Model Fusing} for long document classification while comparing the results with well-known BERT and Longformer architectures. 
\end{abstract}

\section{Introduction}





Text classification is one of the critical tasks in Natural Language Processing, which refers to finding the suitable label/ labels to a particular input text \cite{info10040150, MIRONCZUK201836}. It has a wide range of applications in different domains such as sentiment analysis \cite{electronics9030483,dang2020sentiment}, fake news detection \cite{thota2018fake,kumar2020fake,ahmad2020fake} and offensive language identification \cite{ranasinghe-zampieri-2020-multilingual,husain2021survey}. These tasks are generally referred to as sentence classification tasks since the input text is typically in the form of sentences. In recent years, transformer models such as BERT have provided state-of-the-art results in these text classification tasks \cite{ranasinghe2019brums,gaikwad-etal-2021-cross}. 

While most of the text classification tasks are sentence classification, several domains require classifying lengthy texts into labels typically referred to as document classification. Specifically, domains such as legal and medical often contain long documents that need document classification methods \cite{chalkidis-etal-2019-neural, Hettiarachchi2023}. However, adapting the transformer models that produced state-of-the-art results in sentence classification to document classification is challenging \cite{pappagari2019hierarchical}. The most common transformer models, such as BERT \cite{devlin-etal-2019-bert}, have a limitation of 512 tokens in their input layer, which means the tokens in a lengthy document exceeding this limit will be truncated in the tokenisation step.



The limitations outlined above have garnered significant attention from the research community, leading to the exploration of new document classification architectures. One widely adopted approach is to leverage transformer models that can process longer sequences. Notably, the Longformer \cite{Beltagy2020Longformer} and BigBird \cite{zaheer2020big} transformer models have demonstrated exceptional performance in document classification tasks, with the capacity to accommodate up to 4,096 tokens. However, training transformer models that can process longer sequences is a resource-intensive task, and it may not be feasible for less-resourced domains and languages \cite{wagh2021comparative,zhang2022hierarchical}. In an effort to mitigate this challenge, researchers have attempted to adapt existing pre-trained transformer models to accommodate longer sequences. Notably, two such approaches are Hierarchical BERT \cite{10.1007/978-3-030-88942-5_18} and CogLTX \cite{ding2020cogltx}, both of which propose innovative strategies for adapting BERT to long document classification. Following this, we propose a method to adapt BERT-like transformer models to long document classification using \textit{Model Fusion}. While the methods such as Hierarchical BERT \cite{10.1007/978-3-030-88942-5_18} and CogLTX \cite{ding2020cogltx}  mainly focus on tackling long-term dependencies using different attention mechanisms to reduce their computational complexity, we explore a new idea with model fusing to the long document classification task.

\textit{Model Fusion} refers to the idea of combining several fine-tuned models \cite{xu2020social}. The motivation behind using \textit{Model Fusion} is that multiple models can identify different patterns using different parts of their network, and it is possible to merge multiple models into one model, which will be capable of having all information compressed into a single model. To implement this idea, we divide long documents into multiple parts and use these parts to train part-wise models. Finally, we \textit{fuse} all part-wise models to create a single model capable of handling lengthy sequences. Our evaluation of this approach on four popular document classification datasets shows that while our hypothesis is strong, \textit{Model Fusion}  does not improve state-of-the-art document classification. Nonetheless, we report our results with the aim of helping researchers avoid repeating unsuccessful experiments in the future. Furthermore, this paper identifies potential flaws in experimental design, enabling researchers to refine their methods and improve future studies that employ \textit{Model Fusion} in long document classification\footnote{Publishing negative results has also been encouraged with the organisation of workshops such as Workshop on Insights from Negative Results in NLP \url{https://insights-workshop.github.io/}}.

\noindent Our main contributions of the paper are,

\begin{enumerate}

    \item We present the first study in using \textit{Model Fusion} in long document classification. 

    \item We empirically evaluate the proposed approach in four benchmark datasets in document classification and show that the proposed method does not outperform the baselines such as Longformer \cite{Beltagy2020Longformer}. 

    \item We release the code and the model resources freely available to the public\footnote{Code is available  at \url{https://github.com/DamithDR/legal-classification}}. 
\end{enumerate}


The rest of the paper is organised as follows. Section \ref{sec:related_work} highlights the recent work on long document classification and model fusing. Section \ref{sec:data} describes the datasets we used. Section \ref{sec:method} explains data preparation for experiments, sub-model training, model fusing and prediction on test data. Section \ref{sec:results} presents the results and discusses possible problems in the results and ideas for improvements. Section \ref{sec:conclution} summarises our main experimental findings and conclusions.

\section{Related Work}
\label{sec:related_work}
\paragraph{Long Text Classification}
Over the years, researchers have explored various methods to address long text classification, from traditional machine learning approaches such as SVMs \cite{boser1992training} to recent deep learning architectures \cite{dai-etal-2022-revisiting, uyangodage-etal-2021-transformers}. With the emergence of transformers, the researchers focused heavily on adapting transformer models to long text classification. Longformer \cite{Beltagy2020Longformer} is one such method \cite{hettiarachchi-etal-2021-daai}, which is capable of accommodating 4,096 tokens. Longformer’s attention mechanism is a combination of a windowed local-context self-attention, and an end task motivated global attention that encodes inductive bias about the task. Through ablations and controlled trials, they show both attention types are essential – the local attention is primarily used to build contextual representations, while the global attention allows Longformer to build full sequence representations for prediction. As we mentioned before, training a transformer model that supports lengthy inputs is expensive. Therefore, researchers have explored how to use existing pre-trained transformer models in long document classification. 

CogLTX \cite{ding2020cogltx} is a method which proposes an efficient way of processing long documents using two jointly trained BERT \cite{devlin-etal-2019-bert} models to select
key sentences from long documents for various tasks, including text classification. Their idea is that a few key sentences can be sufficient to get an understanding of the overall text, which works for some tasks but not essentially for document classification. \citet{pappagari2019hierarchical} introduced ToBERT, which can process documents of any length using chunking. However, it does not improve performance in many document classification tasks.


\citet{dai-etal-2022-revisiting} provides a revision on transformers' capabilities on long document classification. \citet{park-etal-2022-efficient} shows a performance comparison between Longformer \cite{Beltagy2020Longformer}, CogLTX \cite{ding2020cogltx}, ToBERT \cite{pappagari2019hierarchical} and their novel baselines BERT+TextRank; where they identify the key sentences using TextRank \cite{mihalcea-tarau-2004-textrank} and uses these sentences to fill the 512 tokens of a BERT rather than using the full document as the input. BERT+Random; is a simpler baseline where they use random sentences to fill the 512 tokens. Interestingly they show that for most of the datasets, specific long-text processing methods fail to outperform these simple baselines. \citet{limsopatham-2021-effectively} has experimented with the effective usage of BERT for long document classification by parsing the front part of the document and the rear part of the document separately and experimenting with the results. Despite numerous efforts to address challenges in long document classification, the results still fall short compared to sentence classification, demanding further dedication from the research community.

\paragraph{Model Fusion}

Fusing is applied on different parts and different levels of NLP tasks. \citet{choshen2022fusing} proposes a way to fuse the models to have better pre-trained models. \citet{xiong-etal-2021-fusing} does label fusing via concatenating texts of labels and an original document to be classified with a [SEP] token as
an input, and they use different segment embeddings for
the label texts and the document text.
\citet{lai2023improving} have used Gated Fusing to improve backward compatibility when doing updates of NLP models. Fusing has been employed in multi-model research, too. \citet{khan2020mmft} provides fusing multiple models for visual question answering. 

As fusion has provided excellent results in different tasks, we hypothesise that fusion can be used to solve document classification. As far as we know, this is the first study to use model fusion in long document classification.


\section{Data}
\label{sec:data}

We evaluated our approach with four popular document classification datasets; ECHR \cite{chalkidis2019neural}, ECHR\_Anon \cite{chalkidis2019neural} 20NewsGroups \cite{lang1995newsweeder} and case-2022 \cite{case-2022-challenges}. We describe each of them below.  The distribution of the number of words in each dataset is also shown in Table \ref{tab:dest_table}. 

\paragraph{ECHR \protect \cite{chalkidis2019neural}}
European Court of Human Rights (ECHR) hears allegations that a state has breached
human rights provisions of the European Convention of Human Rights. The dataset contains approx. 11.5k cases from ECHR’s public database. We use the dataset for document-level binary violation tasks; given the facts of a case, the task is to classify whether there has been any human rights violation or not. 

\paragraph{ECHR\_Anon \protect \cite{chalkidis2019neural}}
This dataset contains an anonymised version of the ECHR with demographic data being anonymised. To achieve this, all Named Entities in the text have been replaced with corresponding tags.

\vspace{-2mm}

\paragraph{20NewsGroups \protect \cite{lang1995newsweeder}}
The dataset is composed of 18828 news articles, which are classified into 20 different categories. The goal of this task is to perform multi-class classification to accurately identify the category of each article. To evaluate our model's performance, we reserve 20\% of the data for the test set.

\paragraph{Case-2022 \protect \cite{case-2022-challenges}}
This dataset is from the shared Task on Socio-political and Crisis Events Detection CASE - subtask 1. The task is a document classification to detect whether a news article contains information about a socio-political event or not. The Dataset features 9384 news articles in the training set, and we have utilised 20\% of it as the test set since the gold labels in the test set are not released.  

\begin{table}[ht]
    \centering
    \scalebox{0.8}{
    \begin{tabular}{c|c|c|c}
        \hline
         Dataset & w $<$ 512  & 512 $<$ w $<$ 4096 & w $>$ 4096 \\
         \hline
         ECHR & 16.04 & 69.15 & 14.80 \\
         ECHR\_Anon & 16.07 & 67.69 & 16.24 \\
         20NewsGroups & 86.72 & 12.67 & 0.61 \\
         Case-2022 & 96.27 & 3.73 & 0.00 \\
    \end{tabular}
    }
    \caption{Percentages of distribution of a number of data instances against the word count (w) in the dataset.}
    \label{tab:dest_table}
\end{table}

\begin{figure}[ht]
    \centering
    \includegraphics[scale=0.45]{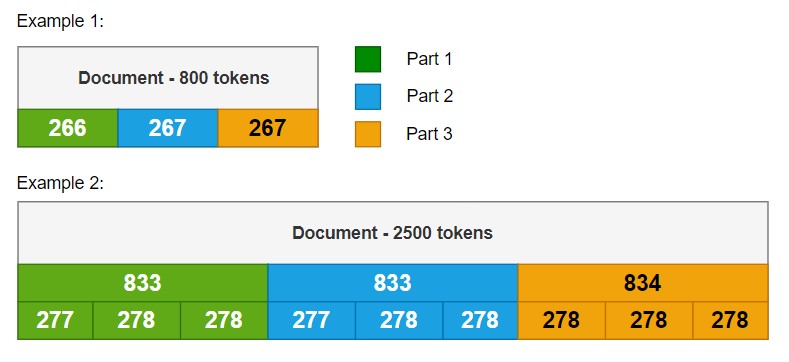}
    \caption{Document breakdown to parts}
    \label{fig:parts_figure}
\end{figure}

\section{Methodology}
\label{sec:method}

We divide our method into five stages, which we describe below. 

\paragraph{\textit{Data Preparation}}
Since the datasets contain data points which exceed 512 token limitation in BERT \cite{devlin-etal-2019-bert} as shown in Table \ref{tab:dest_table}, we evenly distributed each document among sub-models. Initially, we determined the number of parts to divide the data points based on a trial-and-error approach. Early experiments suggested that dividing each data point into three parts produced the best results. We also restricted each part to a maximum of 400 words. For documents with more than 1200 words, such as 3000 words, we split them into three parts of 1000 words each. Due to the 512 token limitation, we further divided the 1000 words into more sub-parts, but all sub-parts were trained on the same model. Essentially, when we split a document into parts, each part has its own respective model that is used for training. To maintain consistency, we assigned respective class labels to the divided parts of the document. We assumed that all parts contribute equally to the class classification, so if the data point had classification label A, all parts of the document would also have the classification label A as illustrated in Figure \ref{fig:parts_figure}.

\begin{figure}[ht]
    \centering
    \includegraphics[scale=0.55]{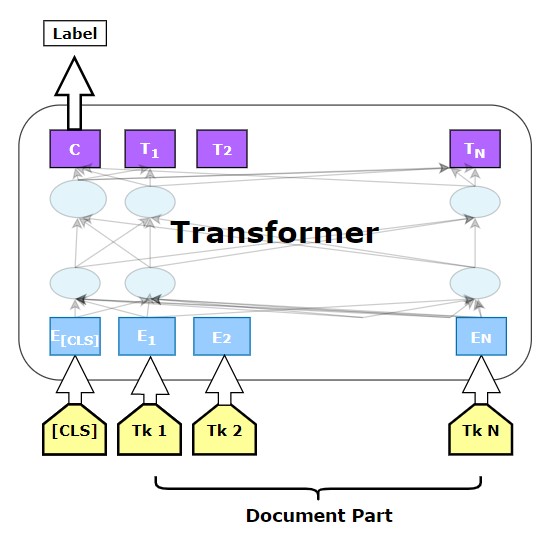}
    \caption{Transformer model for document level classification \cite{uyangodage-etal-2021-multilingual}}
    \label{fig:transformer}
\end{figure}

\begin{figure*}[!ht]
    \centering
    \includegraphics[scale=0.60]{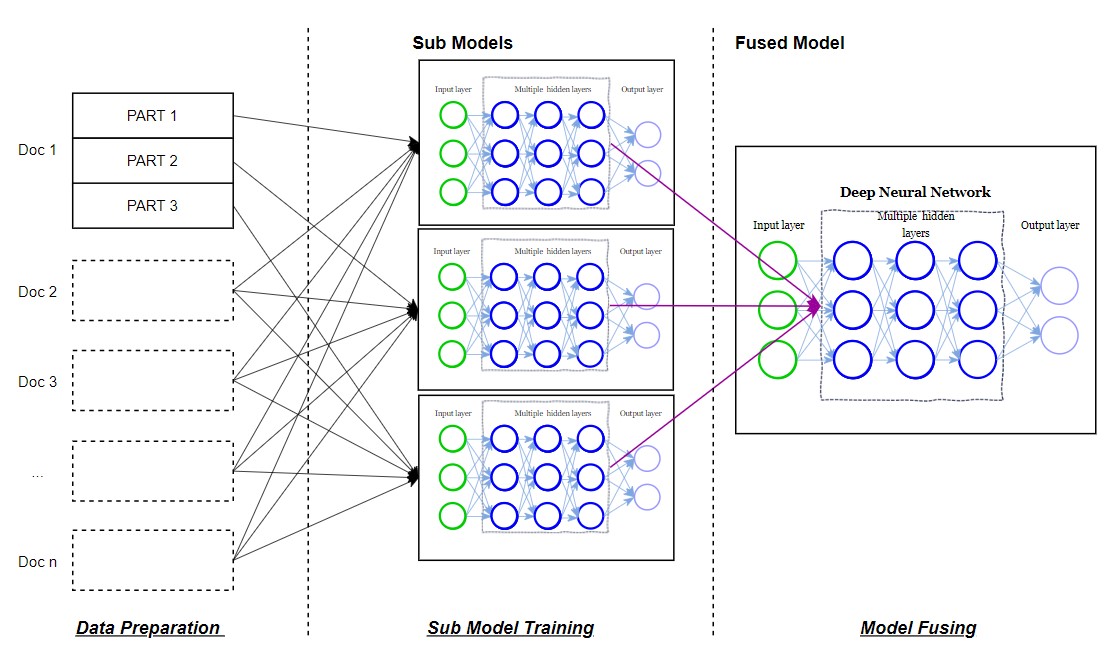}
    \caption{Model fusing pipeline for long document classification}
    \label{fig:fusing_diagram}
\end{figure*}

\paragraph{\textit{Sub-model Training}}
The number of sub-models to be trained is equal to the number of parts in the document. The main idea is to understand the data in a part-localised manner to tackle the length issue. Therefore, in our experiments, we used three sub-models in-line with three parts in each document. As shown in Figure \ref{fig:fusing_diagram}, Part 1 of each document goes to the training set of sub-model 1 and, respectively, part 2 and part 3 into sub-model 2 and 3. We assume that this part-wise modelling can understand the part-local information, which could then contribute to the final classification. Sub-models were trained by using a BERT \cite{devlin-etal-2019-bert} model for all experiments since it has produced excellent results in many natural language processing tasks \cite{morgan-etal-2021-wlv}. We used a softmax layer on top of the last hidden layer of the Transformer architecture, as shown in Figure \ref{fig:transformer}. The configurations we used are listed in Table \ref{tab:train_config}.

\begin{table}[h]
    \centering
    \begin{tabular}{c|c}
        \hline
         Parameter & Value \\
         \hline
         Training Batch Size & 32 \\
         Evaluation Batch Size & 8 \\
         Learning Rate & $4\mathrm{e}{-5}$ \\
         Epochs & 3 \\
         Early Stopping & No \\
         
    \end{tabular}
    \caption{Sub-model training configurations}
    \label{tab:train_config}
\end{table}

\noindent \textbf{\textit{Model Fusing}}
Once the sub-models are trained, we read the weights of hidden layers of the models and fused them together while input and output layers remain unchanged. We employed average fusing for simplicity, in which the resulting fused model has the average of weights in the sub-models as shown in Figure \ref{fig:fusing_diagram}.

\vspace{-3mm}

\begin{equation}
    W_{fused} = f(W_{1},W_{2},...,W_{n}) 
\end{equation}
\begin{equation}
    W_{fused} = (W_{1} + W_{2} + ... + W_{n})/n
\end{equation}

By averaging the weights, we assume that the characteristics of each part of the document are being merged into one fused model. 

\noindent \textbf{\textit{Further Fine-tuning}}
we further fine-tune the fused model using a fraction of the training set, which was split from the training set in the beginning. This step is important as once we merge the models together, the weights of hidden layers are not finely coupled with the output layers. In order to correct this, further fine-tuning step is important and performed using all parts of the document. For this reason, further fine-tune data contain text from all parts separately. In the fine-tune step, we used the same configurations as sub-model training having batch-size of 32, Adam optimiser with learning rate $4\mathrm{e}{-5}$. Once we complete this, the fused model is ready to predict on the test data. 

\noindent \textbf{\textit{Prediction}}
Predicting on test data uses a similar approach to training. We divide the original documents into parts and then predict the classification class for each one of them. We then get the mean of the probabilities of each class and decide the final classification class. We also experimented with taking the max of the probabilities; however, it did not show improvements compared to taking the mean. Therefore, all the results we present were taken using the mean.   


\section{Results and Discussion}
\label{sec:results}
\begin{table*}[ht]
    \centering
    \scalebox{0.9}{
    \begin{tabular}{c|c c c|c c c|c c c}
         Dataset & \multicolumn{3}{c}{\bf Fusing} & \multicolumn{3}{c}{\bf Bert} & \multicolumn{3}{c}{\bf Longformer}  \\
           & P & R & F1 & P & R & F1 & P & R & F1 \\
           \hline
            ECHR & 0.6127 & 0.6451 & 0.5486 & 0.8493 & 0.8486 & 0.8212 & 0.8504 & 0.8516 & \textbf{0.8278}\\
            ECHR\_Anon & 0.6232 & 0.6621 & 0.4673 & 0.8209 & 0.8235 & 0.7950 & 0.8395 & 0.8369 & \textbf{0.8041}\\
            20NewsGroups & 0.5361 & 0.5409 & 0.4984 & 0.8952 & 0.8941 & 0.8910 & 0.8981 & 0.8980 & \textbf{0.8951}\\
            Case-2022 &0.6272 & 0.7920 & 0.4420 & 0.8837 & 0.8858 & 0.8231 & 0.8956 & 0.8981 & \textbf{0.8405}\\
            \hline
           
    \end{tabular}
    }
    \caption{Results for different datasets for Fusing, Bert and Longformer. P; weighted Precision, R; weighted Recall, F1; Macro F1}
    \label{tab:results_table}
\end{table*}

\paragraph{Baselines}
Baseline results were taken from well-known BERT \cite{devlin-etal-2019-bert} and Longformer \cite{Beltagy2020Longformer} which were configured to truncate the sequences which exceeded their token limit. Additionally, Longformer \cite{Beltagy2020Longformer} has the special capability to accommodate up to 4096 tokens.  

\paragraph{Results}
Table \ref{tab:results_table} shows the results for Fusing, BERT and Longformer. It is clear that Longformer performs best among all datasets confirming its unique ability on long document classification. BERT also shows good performance in all cases, and it is clear that 20NewsGroups and Case-2022 datasets are fairly within the range of no of tokens which BERT could capture (512) (Table \ref{tab:dest_table}). However, BERT also performs well in ECHR cases. We believe the reason for that is the first parts of the facts of ECHR cases heavily contribute to the final label.

Fusing results are the lowest in all cases, confirming that model fusing will not produce better results for the long document classification task. It is noticeable that Fusing also has similar trends across datasets as Longformers. Longformer has produced F1 scores of 0.8278 and 0.8041 for ECHR and ECHR\_Anon data, respectively, while Fusing also shows a similar pattern by marking 0.5486 and 0.4673 F1 scores for the same. 

One possible reason for the low performance of the Fusing method could be our assumption where we assumed that all parts of the document equally contribute to its class. This could not be the case at all times, and if not, models will learn incorrect information, which could lead to lower results. Another possibility is the division of the documents into parts. Dividing the documents into parts will induce information flow breaks from which the models could suffer. 

Even though our intuition of model fusing is similar to transfer learning, average fusing has its own problems. Averaging weights might not be ideal because the activation of the neurons could catch with heavy negation. If we average the values 4 and 5, the result is 4.5, which shows that the resulting weight does not deviate from both original weights drastically. However, if we consider 5 and 0.1, their average result is 2.55, which shows a considerable difference between both initial weights. In a numerical model such as BERT, this could introduce significant changes in the network's decision-making process. One way to overcome this issue could be introducing a weighted bias to the sub-models. This way, one model will get favouritism over others and possibly lead to better results, but it will need extensive experiments to confirm this.

\section{Conclusion}
\label{sec:conclution}

This paper presents an empirical study on the effectiveness of model fusing in long document classification, with the aim of comparing its performance to that of state-of-the-art models such as Longformer \cite{Beltagy2020Longformer}. Our results indicate that Longformer \cite{Beltagy2020Longformer} outperforms our experimental setup across all datasets. While we identify several drawbacks of the method, we believe that there is still potential for further exploration in this area. Although our average fusing approach did not yield improved performance in long document classification, there is a need for more research on different fusing methods and their efficacy in various tasks.

\section*{Acknowledgments} 

 We thank the anonymous RANLP reviewers who have provided us with constructive feedback to improve the quality of this paper. 

 We also thank the creators of the datasets we used in the study for making them public. 

\bibliographystyle{acl_natbib}
\bibliography{anthology,ranlp2023}


\end{document}